\documentclass[conference,9pt]{IEEEtran}

\usepackage[T1]{fontenc}
\usepackage{amssymb}	
\usepackage[pdftex]{graphicx}	
\usepackage{xcolor}
\usepackage{mathtools}
\usepackage{amsmath}
\usepackage{algorithmic}
\usepackage[lined,ruled,linesnumbered]{algorithm2e}
\usepackage{cite,balance}

\newcommand{\be}{\boldsymbol{e}}

\newcommand{\bH}{\boldsymbol{H}}

\newcommand{\bJa}{\boldsymbol{\mathcal{J}}}
\newcommand{\bm}{\boldsymbol{m}}

\newcommand{\bn}{\boldsymbol{n}}

\newcommand{\bp}{\boldsymbol{p}}
\newcommand{\bPi}{\boldsymbol{\Pi}}

\newcommand{\bs}{\boldsymbol{s}}
\newcommand{\bt}{\boldsymbol{t}}
\newcommand{\bT}{\boldsymbol{T}}

\newcommand{\bu}{\boldsymbol{u}}
\newcommand{\bmU}{\boldsymbol{\mathcal{U}}}
\newcommand{\bV}{\boldsymbol{V}}
\newcommand{\bv}{\boldsymbol{v}}
\newcommand{\bW}{\boldsymbol{W}}
\newcommand{\bx}{\boldsymbol{x}}
\newcommand{\bX}{\boldsymbol{X}}

\newcommand{\by}{\boldsymbol{y}}

\IEEEoverridecommandlockouts 

\newcommand{\conftitle}{Submitted to EUSIPCO 2017}
\usepackage{fancyhdr}
\fancypagestyle{pageStyleOne}{%
    \fancyhf{}
    \fancyhead[C]{\conftitle}
}
 
\begin{document}
\title{
Multi-frequency image reconstruction for radio-interferometry with self-tuned regularization parameters
}
\author{
\IEEEauthorblockN{Rita Ammanouil, André Ferrari, Rémi Flamary, Chiara Ferrari, and David Mary} 
\IEEEauthorblockA{Universit\'e Côte d'Azur, Observatoire de la Côte d'Azur, CNRS, Lab. J.-L. Lagrange, France} 
\thanks{This work was partly supported by the Agence Nationale pour la Recherche, France, (MAGELLAN project, ANR-14-CE23-0004-01).}
}
\maketitle

\thispagestyle{pageStyleOne}

\begin{abstract}
As the world's largest radio telescope, the Square Kilometer Array (SKA) will provide radio interferometric data with unprecedented detail. 
Image reconstruction algorithms for radio interferometry are challenged to scale well with TeraByte image sizes never seen before. 
In this work, we investigate one such 3D image reconstruction algorithm known as MUFFIN (MUlti-Frequency image reconstruction For radio INterferometry).
In particular, we focus on the challenging task of automatically finding the optimal regularization parameter values. 
In practice, finding the regularization parameters using classical grid search is computationally intensive and nontrivial due to the lack of ground-truth.
We adopt a greedy strategy where, at each iteration, the optimal parameters are found by minimizing the predicted Stein unbiased risk estimate (PSURE). 
The proposed self-tuned version of MUFFIN involves parallel and computationally efficient steps, and scales well with large-scale data.  
Finally, numerical results on a 3D image are presented to showcase the performance of the proposed approach. 
\end{abstract}

\section{Introduction}
\label{intro}

The Square Kilometer Array (SKA) is an international project intended to build the largest radio telescope in the world. 
This telescope will produce interferometric data by combining signals from each pair of antennas in an array measuring radio emissions from a given area in the sky \cite{thompson2008interferometry}. 
With up to one million receiving elements, the SKA will achieve (sub-) arcsec spatial resolution and provide measurements over hundreds of frequency bands  \cite{dewdney2013ska1,dewdney2009square}. 
When operational, it  is expected that the SKA will produce unprecedented images of the sky in the lowest-frequency part of the electromagnetic spectrum. 
However, this will come at the cost of a tremendous volume with data sets sizing up to several hundreds of Pbytes/yr. 
As a result, exploiting these massive data sets in a manageable amount of time will pose considerable challenges \cite{van2013signal}. 

In radio interferometry, most of the data exploitation and analysis relies on producing a proper image of the sky from incomplete Fourier measurements taken by the telescope. 
For perfectly calibrated measurements, this reduces to a deconvolution problem in the image plane aimed at recovering the sky image given the dirty image (the convolved image) and the dirty beam (the point spread function). 
When directional dependent effects and anisoplanetic effects are only partially corrected, the deconvolution takes place in the so-called ``minor loop '' of the image reconstruction pipeline \cite{Jongerius}.
The CLEAN method \cite{hogbom1974aperture} and its multi-frequency extension \cite{rich2008multi} are among the most popular methods for deconvolution. 
CLEAN assumes that the sky is mostly empty, and contains a few point sources. 
The desired ``clean'' image is obtained by successive detections and subtractions of the brightest sources in the dirty image. 
However, CLEAN and most of the algorithms that belong to the CLEAN family are not tailored for large-scale data. 
For this reason, a lot of effort has gone into designing scalable methods based on convex optimization  \cite{Onose16,onose2016randomised,Purify,ferrari2014distributed,deguignet2016distributed,meiller2016,abdulaziz2016low}. 
These algorithms, which unrestrictedly operate in the image or visibility plane, adopt sparse models in an adequate domain, and perform deconvolution by minimizing a data fidelity term and one or more regularizations.
The scalability of these algorithms comes from the fact that computationally intensive steps are performed in parallel. 
In the mono-frequency case \cite{Onose16,onose2016randomised}, parallel computations rely on splitting the data into multiple blocks and exploiting randomization over data blocks at each iteration.
In the multi-frequency case \cite{deguignet2016distributed,meiller2016}, parallel computations exploit the fact that each node can be assigned computations at a certain wavelength band.
In addition to parallelization, scalability is achieved in \cite{deguignet2016distributed} by avoiding expensive operations like matrix inversions. 
Furthermore, the authors of \cite{meiller2016} reduce the memory load by adopting a fully distributed memory storage architecture. 
Despite their various advantages, the efficiency of convex optimization techniques is usually governed by the choice of appropriate regularization parameters. 
The scope of this work is to discuss a scalable method for setting the optimal parameter values.

A typical strategy for parameter tuning is to do a grid search and chose the parameter that gives the best image.  
However, judging the estimated image quality is a nontrivial and complicated task due to the lack of ground truth.
In addition to this, grid search is very time consuming when dealing with large-scale data. 
The literature on automatic parameter selection provides various quantitative measures for evaluating the quality of an estimated image.
These measures can be broadly classified as those based on the discrepancy principle \cite{karl2000regularization,morozov1966solution}, the L-curve method \cite{reginska1996regularization,hansen1993use}, generalized cross-validation (GCV) \cite{golub1979generalized}, and the Stein Unbiased Risk Estimator (SURE) \cite{stein1981estimation,eldar2009generalized}.
Among these measures, the SURE is the most attractive one in image processing since it provides an unbiased estimate of the Mean Square Error (MSE). 
Furthermore, the principles underlying the SURE have been successfully adopted for image deconvolution \cite{ramani2012regularization, giryes2011projected}. 
This work applies a weighted version of SURE, Predicted SURE (PSURE), proposed in \cite{ramani2012regularization} to the MUFFIN (MUlti-Frequency image reconstruction For radio INterferometry) algorithm \cite{deguignet2016distributed} in a greedy way similarly to \cite{giryes2011projected}. 
This new version of MUFFIN can be seen as a self-tuned extension of the original algorithm. 
We show that the steps required for self-tuning can be performed in parallel and in a memory efficient way. 
To the best of our knowledge, \cite{garsden15} is the only work to propose a self-tuning approach. Their approach was restricted to the case of mono-frequency radio-interferometric  image deconvolution, and consists in setting the regularization parameter proportionally to estimated noise variances. 
However, this strategy is confined to the particular case of an inverse problem with a sparse synthesis prior and solved using a primal algorithm.
In contrast with \cite{garsden15}, the application of PSURE is not limited to MUFFIN and can be extended to most deconvolution algorithms based on convex optimization.

The remainder of this article is organized as follows. Section \ref{sec:model} describes the deconvolution problem solved using MUFFIN. 
Section \ref{sec:PSURE} reviews the PSURE and derives its expression for MUFFIN. 
Section \ref{sec:GreedySURE} explains how this measure can be used to find the optimal regularization parameters and the number of iterations. 
Section \ref{sec:simu} presents the experimental results. 

\section{Problem description}
\label{sec:model}

When all direction dependent effects (ionosphere, antenna gains, $\ldots$) and the non-coplanar baseline effects are correctly compensated, the collection of measured visibilities reduces to a sampled Fourier transform of the sky.
Various algorithms \cite{cornwell2008noncoplanar,offringa2014wsclean} transform the sampled visibilities from the Fourier to the image domain resulting in the dirty image. 
The image reconstruction pipeline generally alternates between major and minor loops, that estimate the measurement operator and deconvolve the dirty image  respectively \cite{Jongerius}. 
The deconvolution step recovers the sky image knowing the dirty image and the convolution operator.
Let $\bx_{\ell}{\bf{^\star}}$ be the column vector collecting the sky intensity image at wavelength $\lambda_{\ell}$, with $\ell=1,\ldots,L$ and $L$  the  number of spectral channels. The dirty image $\by_{\ell}$ at wavelength $\lambda_{\ell}$ is related to the sky  image by:
\begin{equation}
\label{model}
    \by_{\ell} = \bH_{\ell} \bx_{\ell}{\bf{^\star}} + \bn_{\ell},
\end{equation}
where $\bH_\ell$ represents a convolution by the point spread function at $\lambda_\ell$, and $\bn_{\ell}$ is a perturbation vector accounting for noise and modeling error. 
Eq.~(\ref{model}) defines an ill-posed deconvolution problem owing to the partial coverage of the Fourier plane. 
Various convex optimization techniques have been proposed the solve this problem. In what follows, we briefly review the approach adopted in \cite{deguignet2016distributed} where the authors solved this problem in a cost minimization framework, by adding to the data fidelity term a regularization term $f_\text{reg}$ representing prior on $\bx_1{\bf{^\star}},\ldots,\bx_{L}{\bf{^\star}}$. 
Let  $\bx_1,\ldots,\bx_{L}$ denote the corresponding optimization variables at each wavelength, and $\bX$ denote the matrix $ \bX := [\bx_1,\ldots,\bx_{L}]$.
With these notations, the cost function  writes:
\begin{equation} 
\label{reg}
\min_{\bX}\; \sum_{\ell=1}^{L} \frac{1}{2 }  \|\by_{l} - \bH_{\ell} \bx_{\ell} \|^2 + f_\text{reg}(\bX).
\end{equation}
The last term in eq. \eqref{reg} incorporates the positivity constraint, a spatial-spectral sparse analysis prior and has the following form:
\begin{equation}
f_\textrm{reg}(\bX) := \boldsymbol{1}_{\mathbb{R}^+}(\bX) + 
\mu_{s}  \sum_{l=1}^{L}   \| \textbf{W}_s \bx_{\ell} \|_1 +
\mu_{\lambda} \sum_{n=1}^{N} \| \textbf{W}_\lambda \bx^{n} \|_1,	 
\label{RegAll}
\end{equation}
where $\bx^n$ (the $n^{th}$ row of $\bX$) is the spectrum associated to  pixel $n$,  
$\textbf{W}_s$ and $\textbf{W}_\lambda$ are the operators associated  with, respectively, the spatial and spectral decomposition,
$\mu_s$ and $\mu_\lambda$ are the corresponding regularization parameters. 
Without loss of generality, we considered in the experiments a union of orthogonal bases for the spatial regularization and  a cosine decomposition for the spectral model  widely used in the literature \cite{Purify,Onose16, garsden15}. 
The authors of \cite{deguignet2016distributed} solve the optimization problem defined in \eqref{reg} and \eqref{RegAll} using the MUFFIN algorithm based on the primal-dual algorithm proposed in \cite{Condat,Vu11}. 
The steps required for each iteration of MUFFIN are described in Algorithm \ref{algo1} where$(\cdot)_+$  is the projection on the positive orthant, and  $\text{sat}(u)$ is the saturation function defined as follows:
\begin{alignat}{2}
\text{sat}(u) := 
\begin{dcases}
  -1 & \quad \text{if} \quad u < -1\\
  1 & \quad \text{if} \quad u > 1\\ 
  u &  \quad \text{if} \quad |u| \leq 1,
\end{dcases}
\end{alignat}
and the parameters $\sigma$ and $\tau$ are fixed according to \cite{Condat} in order to guarantee convergence. 
A major advantage of the MUFFIN Algorithm is that the most computationally demanding steps are separable w.r.t.
the wavelengths, leading to a parallel and distributed implementation.
More precisely, each wavelength is associated to a compute node where the  time consuming steps are computed in parallel which drastically reduces the execution time.
This is in contrast with the work of \cite{abdulaziz2016low} where the most expensive step requiring a singular value decomposition is not parallelized.



\begin{algorithm}
\SetKwInOut{Input}{Initialize}
\SetKwInOut{Output}{Return}

The master computes $\bT = \mu_\lambda\bV{\bW_\lambda}$. Equivalently, it computes:
\begin{equation}
\bt^n = \mu_\lambda\bW_\lambda^\dagger\bv^n,  \quad \text{for}  n=1\ldots N  \label{t}
\end{equation}
and sends the col. $\ell$ of $\bT$, denoted as $\bt_\ell$, to node $\ell$. \quad \quad \quad \quad \quad \quad \quad \quad \quad \quad

Each compute node $\ell=1\ldots L$ computes sequentially:
\begin{align}
\bx_\ell &=  \tilde{\bx}_\ell ,    \label{updatenode} \\
\boldsymbol{\nabla}_\ell &=\bH_\ell^\dagger(\bH_\ell\bx_\ell-\by_\ell)  \label{grad} \\  
\bs_\ell &=\mu_s\bW_s^\dagger \bu_\ell \label{adjointspat}  \\
\bm_\ell &= \bx_\ell - \tau(\boldsymbol{\nabla}_\ell+\bs_\ell + \bt_\ell)   \\
\tilde{\bx}_\ell &= \left( \bm_\ell  \right)_+   \\
{\bp}_\ell &=\bu_\ell + \sigma\mu_s\bW_s(2\tilde{\bx}_\ell -\bx_\ell)  \\
{\bu}_\ell &=\text{sat}\left({\bp}_\ell\right), \label{updateu}
\end{align}
and sends $\bx_\ell$ and $\tilde{\bx}_\ell$ to the master node. \quad \quad \quad \quad \quad \quad \quad \quad \quad \quad \quad \quad \quad \quad \quad

The master computes sequentially, for $n=1\ldots N$:
\begin{align}
 {\tilde{\bv}}^n &=\bv^n + \sigma\mu_\lambda\bW_\lambda(2\tilde{\bx}^n -\bx^n) \label{vt}   \\
 {\bv}^n &=\text{sat}\left( {\tilde{\bv}}^n \right)   \label{v}
\end{align}

\Output{$\bX:= [\bx_1,\ldots,\bx_{L}]$}
\caption{MUFFIN update \label{algo1}} 
\end{algorithm}

\section{Predicted-SURE measure}
\label{sec:PSURE}

The SURE provides an unbiased estimate of the MSE. However the SURE measure is only applicable in the case of image denoising, which is not the case in the present study. 
The concepts underlying the SURE have been extended to the case of image reconstruction using weighted versions of the SURE  measure  \cite{ramani2012regularization,giryes2011projected}.
In what follows, we briefly review the essential concepts behind one such measure, the PSURE measure, proposed in \cite{ramani2012regularization} and how it can be used to automatically select the optimal parameter values. 
Similarly to the SURE, PSURE provides an unbiased estimate of the weighted MSE that is only a function of the measurements. More precisely: 
\begin{equation}
\sum_{\ell=1}^L \mathbb{E}\left[ \|\bH_\ell(\bx^\star_\ell - \bx_\ell)\|^2 \right] =  \sum_{\ell=1}^L \mathbb{E}\left [ \text{PSURE}_\ell \right ], \\
\label{PSURE1}
\end{equation}
with 
\begin{equation}
\mathsf{PSURE}_\ell {:=}  \|\by_\ell - \bH_\ell \bx_\ell \|^2 + {2\sigma^2_\ell}\text{tr}\left \{ \bH_\ell \bJa(\bx_\ell)\right \} - N\sigma^2_\ell,
\label{PSURE}
\end{equation}
where $\sigma_{l}$ is the noise variance at $\lambda_{\ell}$, and $\bJa(\bx_\ell)$ is the Jacobian matrix of $\bx_\ell$ with respect to the measurements $\by_\ell$ such that the $(i,j)$-th element of the resulting matrix is defined as
\begin{equation}
\left[\bJa(\bx_\ell)\right]_{i,j} = \frac{\partial x_{i,\ell}}{\partial y_j}.
\label{Jac}
\end{equation}
Note that all jacobians used in the rest of the article are computed with respect to the measurements, similarly to $\bJa(\bx_\ell)$.
It is important to note that while the PSURE estimates a weighted MSE, being a function of $\bx$, it is also dependent on the set of parameters used to estimate $\bx$ i.e. $\mu_s$ and $\mu_\lambda$. As a result, one may search for the values of $\mu_s$ and $\mu_\lambda$ that minimize the PSURE, and in that way aim at minimizing the true weighted MSE. 
The estimation of the PSURE in \eqref{PSURE}, requires the knowledge of the noise variance and the estimation of the Jacobian matrices $\bJa(\bx_\ell)$ at each band. 
Note that, $\bx_\ell$ does not have a closed form expression as a function of $\by_\ell$, as a result, it is not possible to have a closed form expression for the Jacobian matrices. 
One alternative is to estimate these matrices iteratively according to the estimation of $\bx_l$ in algorithm \ref{algo1}. 
We derived the Jacobian matrix expression of MUFFIN's reconstruction operator with respect to the measurements. 
This is done using  straightforward applications of the product and the chain rules for derivations. 
The derivatives are interpreted in the weak sense of distributions whenever the function is not differentiable. 
Nevertheless, one concern with the Jacobians estimation is that these matrices have enormous sizes. 
According to the definition in \eqref{Jac}, the number of columns of $\bJa(\bx_\ell)$ equals the size of the data cube.
For a typical reconstruction setting in radio-interferomery, it would be impossible to store or manipulate this matrix. 
In order to avoid the storage of Jacobian matrices, we use the following stochastic approximation \cite{hutchinson1990stochastic} that provides an estimate of the trace required in equation \eqref{PSURE}: 
\begin{equation}
\text{tr}\{\bH_\ell \bJa(\bx_\ell)\} \approx \be^t\bH_\ell \bJa(\bx_\ell)\be
\label{approx}
\end{equation}
where $\be$ an i.i.d. zero-mean random vector with unit variance.
As a result, in order to estimate the last term  in equation \eqref{PSURE}, we can iterate over
\begin{equation}
\bJa_e(\bx_\ell) := \bJa(\bx_\ell) \be
\end{equation}
rather than $\bJa(\bx_\ell)$, which is the size of $\bx$. 
Using this approximation, the PSURE estimation for the MUFFIN algorithm can be updated at each iteration of algorithm \ref{algo1} according to the steps  described in algorithm \ref{algo2}.
In the algorithm, we used the notation $\text{Diag}(\bu)$ to refer to the diagonal function which returns a square diagonal matrix with the elements of its input vector $\bu$ on the main diagonal. The functions $\mathcal{U}(\cdot)$ and $\Pi(\cdot)$ are applied component-wised and defined as:
\begin{alignat}{2}
\mathcal{U}(u) := 
\begin{dcases}
  0 & \quad \text{if} \quad u \leq 0 \\
  1 & \quad \text{if} \quad u > 0 \\ 
\end{dcases}
\end{alignat}
and 
\begin{alignat}{2}
\Pi(u) := 
\begin{dcases}
  1 & \quad \text{if} \quad -1 \leq u \leq 1 \\
  0 & \quad \text{if} \quad \text{otherwise}. \\ 
\end{dcases}
\end{alignat}

\section{Automated parameter selection}
\label{sec:GreedySURE}

The global method for setting one of the regularization parameters is to find the minimizer of the PSURE at the convergence of algorithm \ref{algo1}, using for example a golden section search.  
We adopt the greedy approach proposed in \cite{giryes2011projected} where the parameter is updated at each iteration. More precisely, a golden section search is performed at each iteration of the algorithm. In this case, the algorithm can be stopped when the value of the estimated weighted MSE, i.e. the PSURE, is smaller than a predefined threshold. 
This way the number of iterations is automatically set together with the regularization parameter. 

In MUFFIN, two regularization parameters have to be set, namely $\mu_s$ and $\mu_\lambda$.
We adopt a greedy strategy described in algorithm \ref{algo3} which decouples the calibration in two steps, and finally runs MUFFIN with the fixed optimal parameters. 
In the first set of iterations,  $\mu_\lambda$ is set to $0$ and PSURE is used to automatically tune $\mu_s$. 
Note that in this case the problem is separable w.r.t. the wavelengths, and each node independently iterates Eqs.~\eqref{updatenode}-\eqref{updateu} and $\bt_\ell = \boldsymbol{0}$. 
This setting reduces the computational load on the master node. However, the master node still receives the PSURE estimate at each band and centralizes the computation of the overall PSURE estimate in order to choose the best regularization parameter. 
In the second set of iterations, $\mu_s$ is fixed to its optimal value (from previous step) and the automatic tuning of $\mu_\lambda$ is triggered using the same scheme as before.
In the last step MUFFIN is ran with the optimal fixed parameters allowing to ensure and speed-up the convergence.

\begin{algorithm}
\SetKwInOut{Input}{Initialize}
\SetKwInOut{Output}{Return}

The master computes $\bJa(\bT)$. Equivalently, it computes:
\begin{equation}
\bJa_e( \bt^n) = \mu_\lambda\bW_\lambda^\dagger  \bJa_e( \bv^n),  \quad \text{for}  n=1\ldots N
\end{equation}
and sends the col. $\ell$ of $\bJa_e(\bT)$, denoted as $\bJa_e(\bt_\ell)$, to node $\ell$.  \quad \quad \quad 

Each compute node $\ell=1\ldots L$ computes sequentially:
\begin{align}
\bJa_e( \bx_\ell ) & =  \bJa_e(\tilde{\bx}_\ell) \\
\bJa_e( \boldsymbol{\nabla}_\ell) & =\bH_\ell^\dagger(\bH_\ell  \bJa_e(  \bx_\ell ) -\by_\ell)  \\
\bJa_e( \bs_\ell) & =\mu_s\bW_s^\dagger  \bJa_e( \bu_\ell)   \\
\bJa_e( \bm_\ell)  & = \bx_\ell - \tau(  \bJa_e(  \boldsymbol{\nabla}_\ell) +  \bJa_e( \bs_\ell ) +  \bJa_e( \bt_\ell))   \\
\bJa_e( \tilde{\bx}_\ell) &  = \text{Diag}\left\{\bmU( \bm_\ell  )\right\} \bJa_e(\bm_\ell )  \\
\bJa_e({\bp}_\ell) & =\bu_\ell + \sigma\mu_s\bW_s(2\bJa_e(\tilde{\bx}_\ell) - \bJa_e(\bx_\ell))  \\
\bJa_e( {\bu}_\ell) & =  \text{Diag}\left\{\bPi\left( {{\bp}}_\ell \right)\right\} \bJa_e({{\bp}}_\ell),     
\end{align}
and sends $ \bJa_e(\bx_\ell)$ and $ \bJa_e(\tilde{\bx}_\ell)$ to the master. \quad \quad \quad \quad \quad \quad \quad \quad \quad \quad \quad \quad \quad \quad \quad

The master computes sequentially, for $n=1\ldots N$:
\begin{align}
& \bJa_e(  {\tilde{\bv}}^n)= \bJa_e( \bv^n) + \sigma\mu_\lambda\bW_\lambda(2  \bJa( \tilde{\bx}^n ) - \bJa( \bx^n) ) \nonumber \\
& \bJa_e(  {\bv}^n)=  \text{Diag}\left\{\bPi\left( {\tilde{\bv}}^n \right)\right\} \bJa_e({\tilde{\bv}}^n),   
\end{align}
and estimates PSURE$=\sum_{\ell=1}^L \text{PSURE}_\ell$ eq. \eqref{PSURE1}--\eqref{approx}

\Output{PSURE}
\caption{PSURE update \label{algo2}}
\end{algorithm}

\begin{algorithm}
\SetKwInOut{Input}{Initialize}
\SetKwInOut{Output}{Return}

$\mu_\lambda=0$

\Repeat{ stopping criteria $1$}{
MUFFIN update (alg. \ref{algo1})

$\mu_s$ = $\text{argmin}_{\mu_s} \text{PSURE}$ (alg. \ref{algo2} + golden section)
}

\Repeat{ stopping criteria $2$}{
MUFFIN update (alg. \ref{algo1})

$\mu_\lambda$ = $\text{argmin}_{\mu_\lambda} \text{PSURE}$ (alg. \ref{algo2} + golden section)
}

\Repeat{ stopping criteria $3$}{
MUFFIN update (alg. \ref{algo1})
}
\Output{$\bX$}
\caption{Self-tuned MUFFIN \label{algo3}}
\end{algorithm}

\section{Simulations}
\label{sec:simu}
\begin{figure}[]
\center
\begin{minipage}[b]{0.9\linewidth}
  \centering
  \centerline{
  \includegraphics[trim =  .5cm 1.5cm 0cm 0.8cm, clip = true,width=0.99\textwidth,height=0.99\textheight,keepaspectratio]{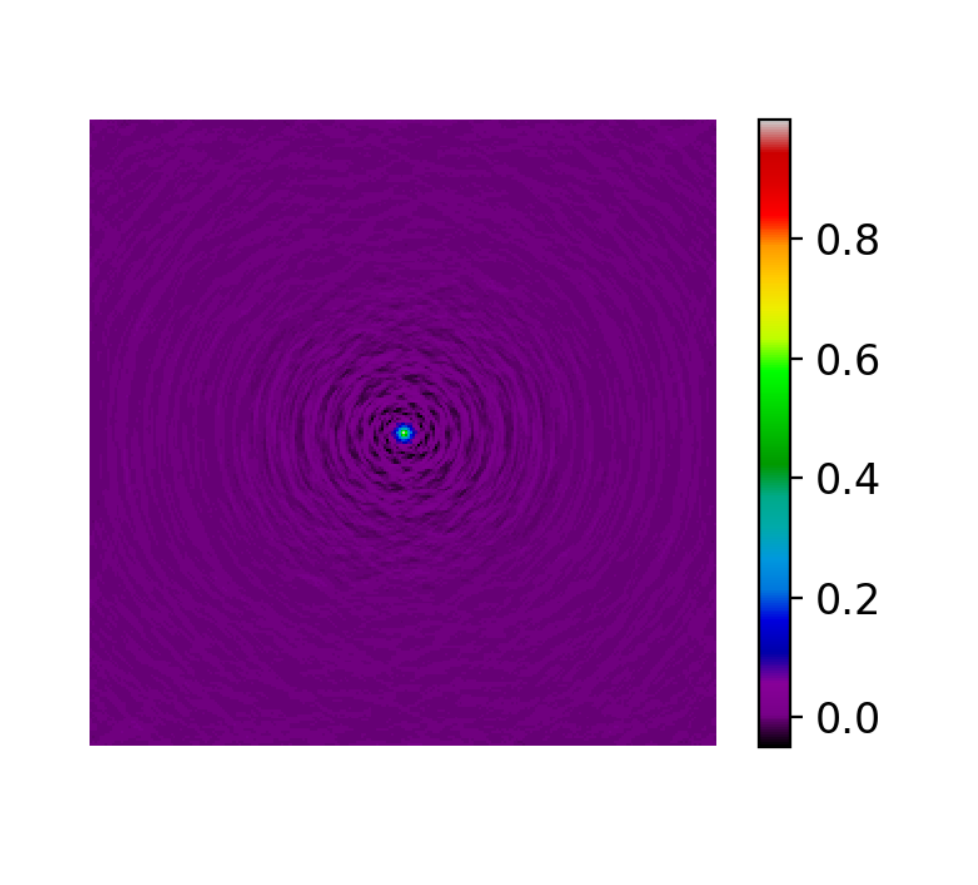}
  }
\end{minipage}
\caption{Point spread function (PSF) at the $100$-th wavelength band showing the central lobe and the ringed side-lobes at larger angular distances.}
\label{fig:PSF}
\end{figure}

We simulate the PSF of MeerKAT\footnote{MeerKAT is the South African SKA pathfinder instrument.} observations using the HI-inator package \footnote{\texttt{https://github.com/SpheMakh/HI-Inator}} based on the MeqTrees software \cite{noordam2010meqtrees}. 
We simulated a cube with $100$ frequency bands and a size of $256\times 256$ pixels.
Figure \ref{fig:PSF} shows the PSF at the $100$-th wavelength band which corresponds to a Fourier coverage produced by a total observation time of $8$ hours.
The sky corresponds to the radio emission of an HII region in the M$31$ galaxy.  
The sky cube for this real sky image is computed using first order power-law spectrum models. 
The $256\times 256$ map of spectral indices is constructed following the procedure detailed in \cite{junklewitz2015new}: for each pixel, the spectral index is a linear combination of an homogeneous Gaussian field and
the reference sky image.
For $\textbf{W}_s$ and $\textbf{W}_\lambda$, we considered a union of $8$ Daubechies wavelet bases for the spatial regularization and  a cosine decomposition for the spectral model.
The dirty image was created by convolving the sky image with the PSF, and Gaussian noise was finally added to the dirty image such as the resulting signal to noise (SNR) is equal to $10$ dB. The parameters of the algorithm are set to $\sigma=10$ and $\tau=10^{-3}$.
The simulations were performed using $20$ compute nodes\footnote{This work was granted access to the HPC and visualization resources of ``Centre de Calcul Interactif'' hosted by ``Université Nice Sophia Antipolis''} which corresponds to $5$ spectral bands per node. The algorithm is written using Python, and the Message Passing Interface (MPI) is used to deal with the parallel computing architecture.  

\begin{figure}[]
\center
\begin{minipage}[b]{0.801\linewidth}
  \centering
  \centerline{
  \includegraphics[trim =  .5cm .1cm .8cm .1cm, clip = true,width=0.99\textwidth,height=0.99\textheight,keepaspectratio]{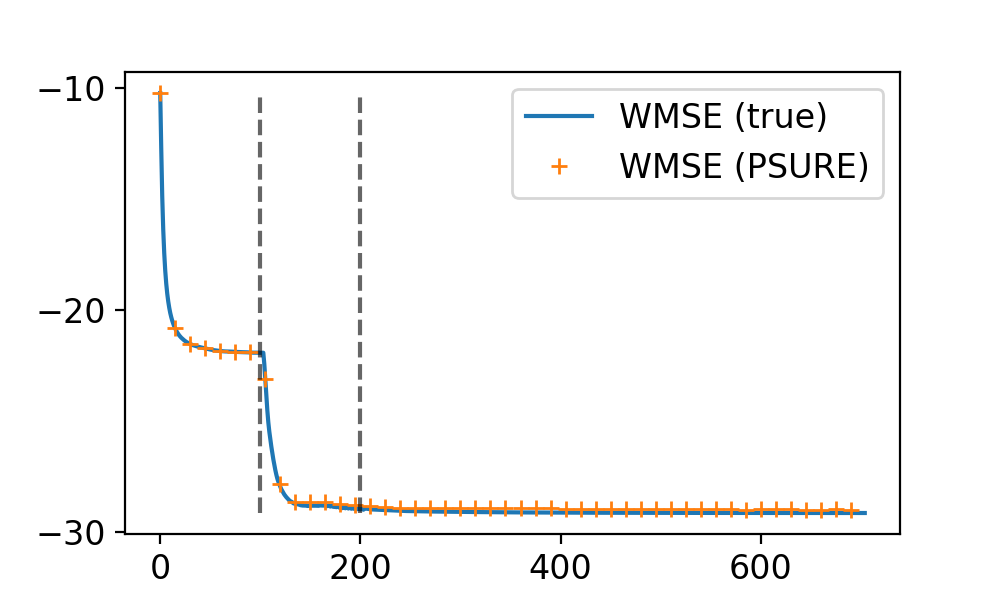}
  }
\end{minipage}
\caption{Variation of the true WMSE and $\widehat{\text{WMSE}}$ (in dB) with the number of iteration obtained with self-tuned MUFFIN. Step $1$ (from iter. $0$--$100$): $\mu_\lambda=0$ and $\mu_s$ is updated by the algorithm. Step $2$ (from iter. $101$--$200$): $\mu_s=0.27$ and $\mu_\lambda$ is updated by the algorithm. Step $3$ (from iter. $201$--$700$): $\mu_s=0.27$ and $\mu_\lambda=1.93$.}
\label{fig:wmse}
\end{figure}

We tested self-tuned MUFFIN on the simulated data set. The golden section search intervals for $\mu_s$ and $\mu_\lambda$ were set to $[0, ~ 2]$ and $[0, ~ 3]$ respectively. In the first step, $\mu_\lambda$ is set to $0$, and self-tuned MUFFIN automatically found that the optimal $\mu_s$ is equal to $0.27$. In the second step, $\mu_s$ is set to its optimal value $0.27$, self-tuned MUFFIN automatically found that the optimal $\mu_\lambda$ is equal to $1.93$. Figure \ref{fig:wmse} shows the variation of the true weighted mean square error (WMSE): 
\begin{equation}
\text{WMSE}(\bX,\bX^{\bf{\star}}) = \frac{1}{LN} \sum_{\ell=1}^L\|\bH_\ell(\bx_\ell - \bx_\ell^\star)\|_{\text{F}}^2,
\end{equation}
and the WMSE estimated using PSURE, which is the one actually used by self-tuned MUFFIN, and defined as
\begin{equation}
\widehat{\text{WMSE}}(\bX,\bX^{\bf{\star}}) = \frac{1}{LN} \sum_{\ell=1}^L   \text{PSURE}_\ell.
\end{equation}
Note that the true WMSE and the WMSE estimated using PSURE are reported in dB. The dashed gray lines in the Figure delimit step $1$ (from iter. $1$--$100$), steps $2$ (from iter. $101$--$200$), and step $3$ (from iter. $201$--$700$). 
More precisely, in the first step $\mu_\lambda=0$ and $\mu_s$ is updated, while in the second step $\mu_s=0.27$ and $\mu_\lambda$ is updated by the algorithm, and finally in the third step $\mu_s=0.27$ and $\mu_\lambda=1.93$.
Note that we could have stopped after the first $200$ iterations, nevertheless step $3$ allows to verify that the algorithm has definitely converged. 
Furthermore each iteration in step $3$ is relatively fast since it does not involve a golden section search.
Figure \ref{fig:wmse} shows that the PSURE estimate is a very good estimate of the WMSE. In addition to this, it shows that the WMSE decreased throughout the iterations, and it had an abrupt decrease when the spectral regularizer was triggered. 

\begin{figure}[]
\center
\begin{minipage}[b]{0.801\linewidth}
  \centering
  \centerline{
  \includegraphics[trim =   .5cm .1cm .8cm .1cm, clip = true,width=0.99\textwidth,height=0.99\textheight,keepaspectratio]{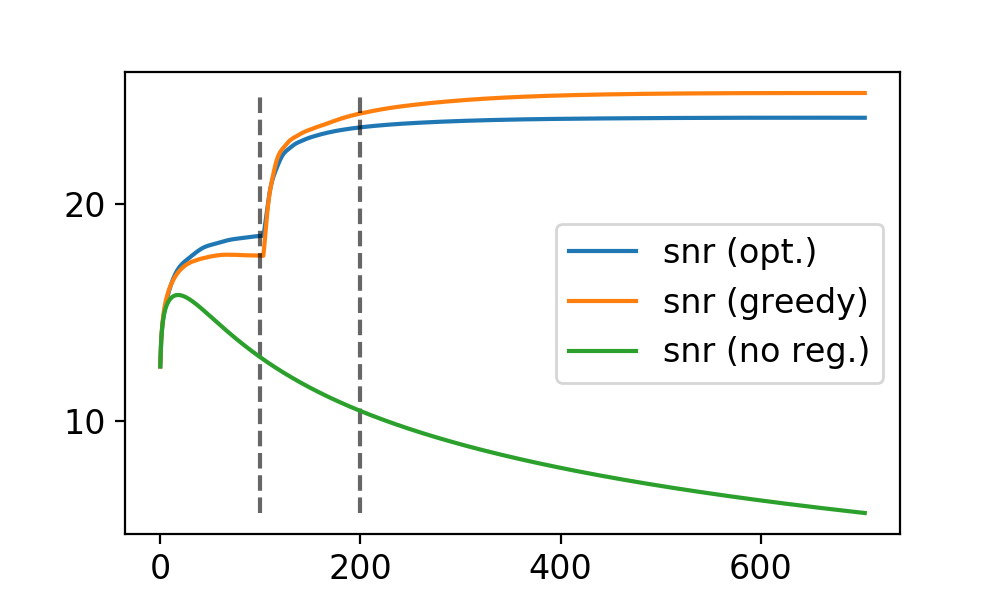}
  }
\end{minipage}
\caption{Variation of the SNR (in dB) as a function of the iterations obtained with self-tuned MUFFIN using the parameters found by the greedy golden section search (greedy) $\mu_s=0.27$ and $\mu_\lambda=1.93$, with MUFFIN using the optimal parameters found by the golden section search $\mu_s=0.43$ and $\mu_\lambda=2.20$, and with MUFFIN using $\mu_s=0$ and $\mu_\lambda=0$.}
\label{fig:snr}
\end{figure}

These results were compared to the results obtained using optimal fixed parameters. 
More precisely, we ran each time MUFFIN using a pre-fixed value of $\mu_s$ in the first $100$ iterations ($\mu_\lambda$ being set to zero), and then by setting $\mu_\lambda$ to some pre-fixed value in the next $600$ iterations. 
The number of iterations was set to $100$ in the first step and $600$ in the second step in order to be comparable to the setting used with the self-tuned MUFFIN in the previous simulation. 
We used the golden section search to find the optimal $\mu_s$ that would yield the lowest WMSE at iteration $100$. Then the golden section was used to find the optimal $\mu_\lambda$ that would yield the lowest WMSE at iteration $700$ given that the first $100$ iterations were run with the previously found optimal value for $\mu_s$ and $\mu_\lambda=0$. 
In the golden section search, $\mu_s$ and $\mu_\lambda$ were constrained to the same intervals used previously, $[0 , ~ 2]$ and $[0 , ~ 3]$ respectively.  
This time, we used the real sky to estimate the weighted mean-square error at convergence. 
The golden section search with MUFFIN found $\mu_s=0.43$ and $\mu_\lambda=2.20$. These values are slightly higher than the optimal values found using the greedy search in  self-tuned MUFFIN. 
Figure \ref{fig:snr} shows the variation of the SNR defined as:
\begin{equation}
\textrm{SNR}(\bX,\bX^{\bf{\star}}) := 10 \log_{10}\left(\frac{\| {\bX^{\bf{\star}}} \|_{2}^{2}}{\| {\bX} - \bX^{\bf{\star}}  \|_{2}^{2}}\right), 
\end{equation}
obtained with the self-tuned MUFFIN, MUFFIN using the optimal parameters $\mu_s=0.43$ and $\mu_\lambda=2.20$, and MUFFIN using $\mu_s=0$ and $\mu_\lambda=0$. Figure \ref{fig:snr} shows that both self-tuned MUFFIN and MUFFIN with the optimal parameters outperformed MUFFIN when the regularization parameters where both set to zero. 
It also shows that in the first step, MUFFIN with the optimal $\mu_s$ outperformed self-tuned MUFFIN. However, beyond the first step, self-tuned MUFFIN was able to  catch up very quickly with the optimally tuned MUFFIN and eventually outperformed it. This is due to the adopted two-step tuning strategy which is computationally efficient but sub-optimal compared to a simultaneous optimization of both parameters. As a result, the optimally tuned MUFFIN was over-regularized w.r.t. $\mu_s$ compared to self-tuned MUFFIN. 
Finally, note that the self-tuned MUFFIN code will be available on the authors Github pages. 

\section{Conclusion}
In conclusion, we proposed an extended version of MUFFIN denoted as self-tuned MUFFIN. Self-tuned MUFFIN is endowed with a parallel and computationally efficient implementation, and it allows to automatically find the optimal regularization parameters.

\begin{figure}[]
\begin{minipage}[b]{.32\linewidth}
  \centering
  \centerline{
\includegraphics[trim =  2.0cm 0.5cm 1.0cm 0.90cm , clip = true,width=0.99\textwidth,height=0.99\textheight,keepaspectratio,angle=90]{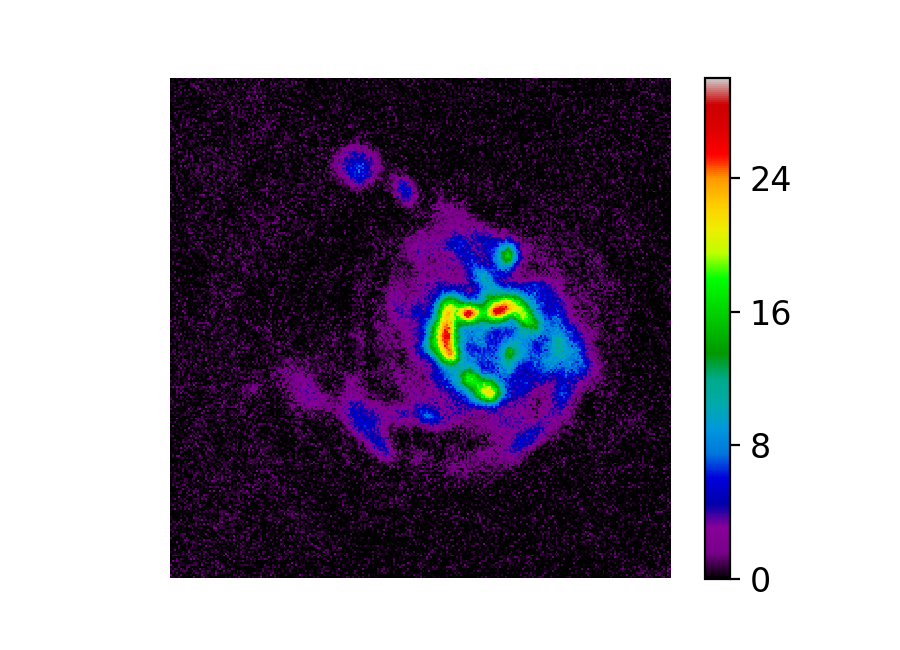} 
  }
\end{minipage}
\begin{minipage}[b]{0.32\linewidth}
  \centering
  \centerline{
   \includegraphics[trim = 2.0cm 0.5cm 1.0cm 0.90cm, clip = true,width=0.99\textwidth,height=0.99\textheight,keepaspectratio,angle=90]{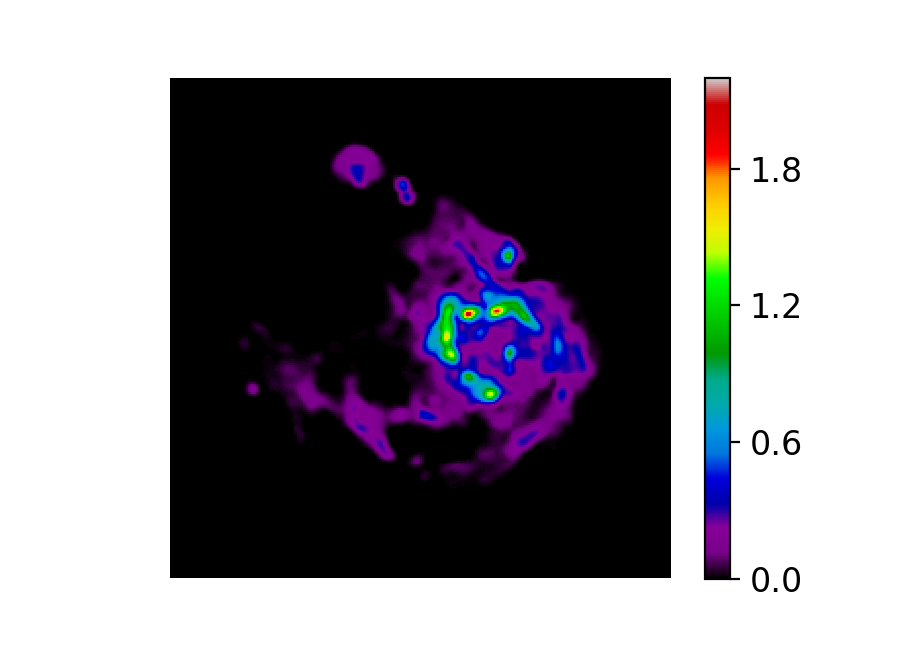}
  }
\end{minipage}
\begin{minipage}[b]{0.32\linewidth}
  \centering
  \centerline{
   \includegraphics[trim = 2.0cm 0.5cm 1.0cm 0.90cm, clip = true,width=0.99\textwidth,height=0.99\textheight,keepaspectratio,angle=90]{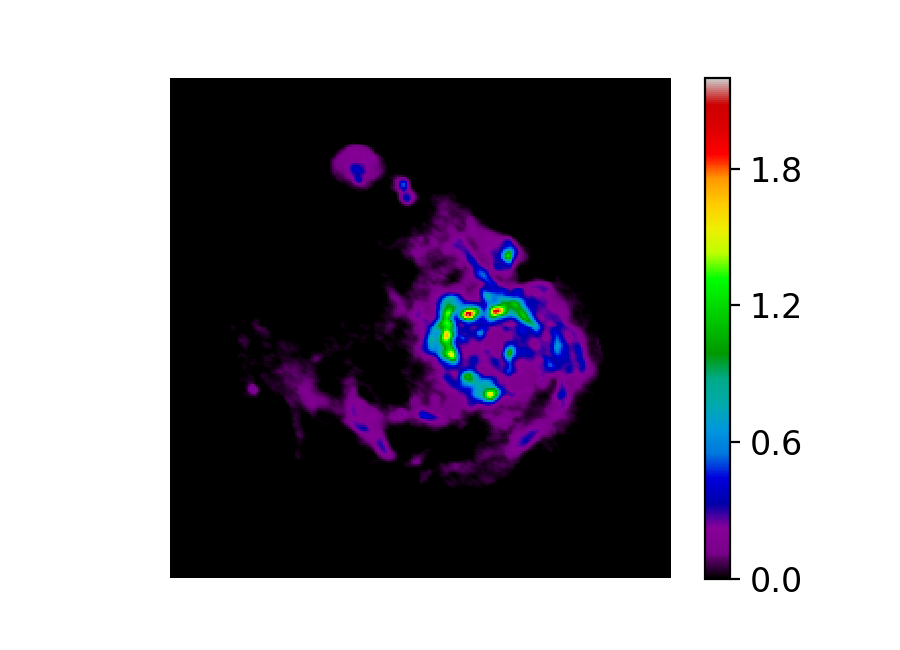}
  }
\end{minipage}


\begin{minipage}[b]{.32\linewidth}
  \centering
  \centerline{
\includegraphics[trim =  2.0cm 0.5cm 1.0cm 0.90cm, clip = true,width=0.99\textwidth,height=0.99\textheight,keepaspectratio,angle=90]{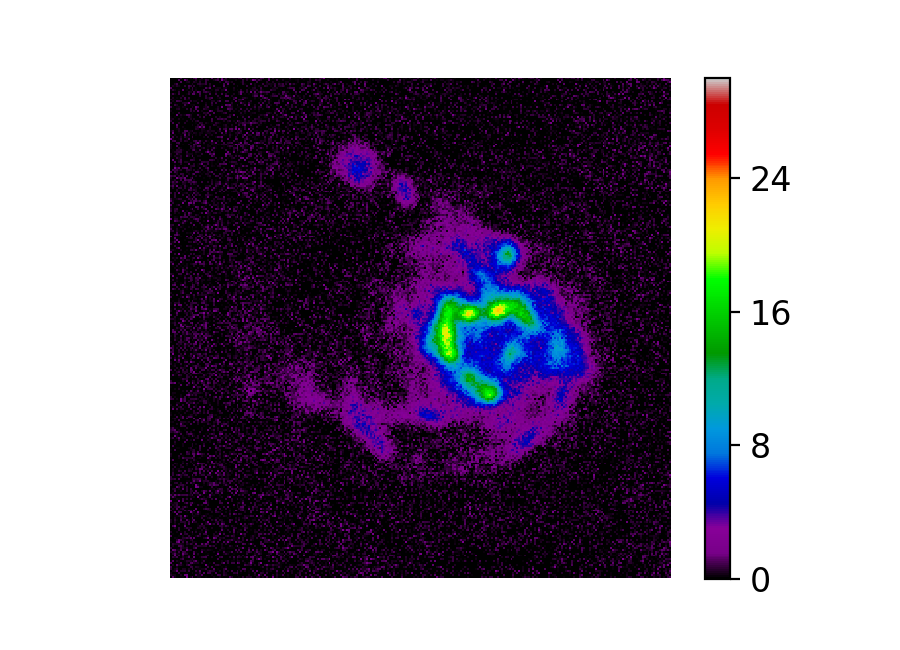} 
  }
\end{minipage}
\begin{minipage}[b]{0.32\linewidth}
  \centering
  \centerline{
   \includegraphics[trim = 2.0cm 0.5cm 1.0cm 0.90cm, clip = true,width=0.99\textwidth,height=0.99\textheight,keepaspectratio,angle=90]{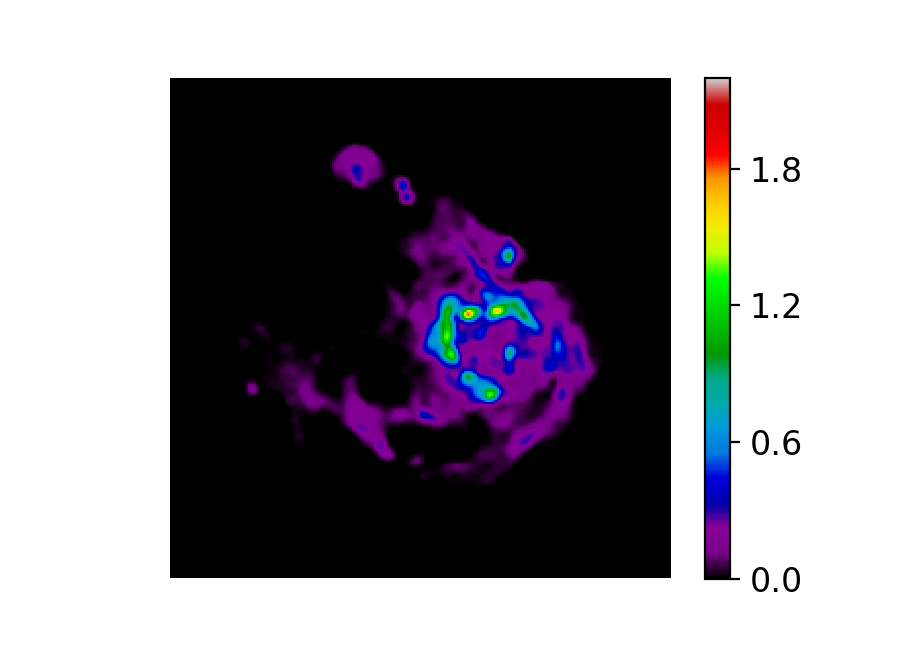}
  }
\end{minipage}
\begin{minipage}[b]{0.32\linewidth}
  \centering
  \centerline{
   \includegraphics[trim = 2.0cm 0.5cm 1.0cm 0.90cm, clip = true,width=0.99\textwidth,height=0.99\textheight,keepaspectratio,angle=90]{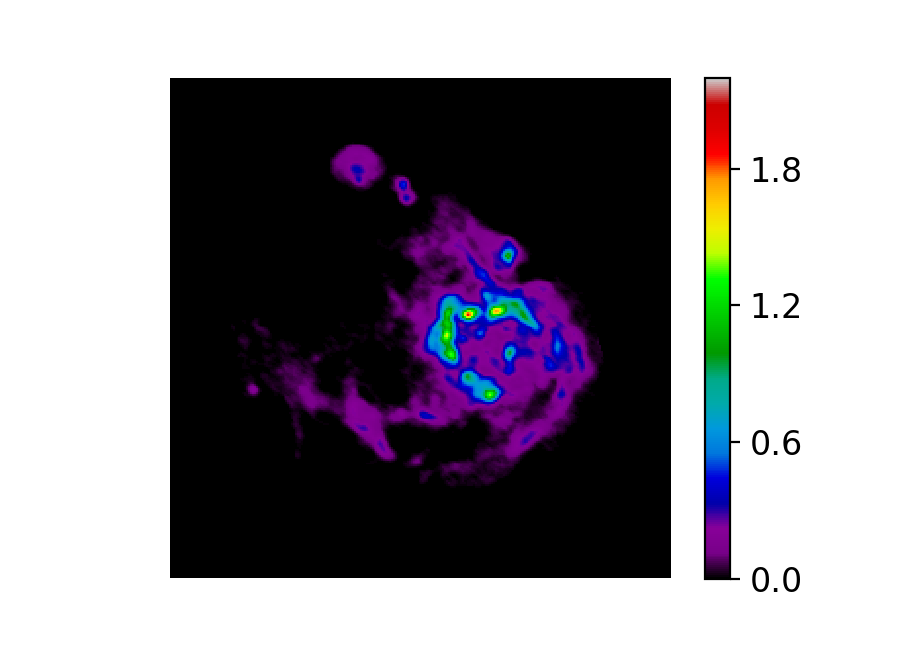}
  }
\end{minipage}


\begin{minipage}[b]{.32\linewidth}
  \centering
  \centerline{
\includegraphics[trim =  2.0cm 0.5cm 1.0cm 0.90cm, clip = true,width=0.99\textwidth,height=0.99\textheight,keepaspectratio,angle=90]{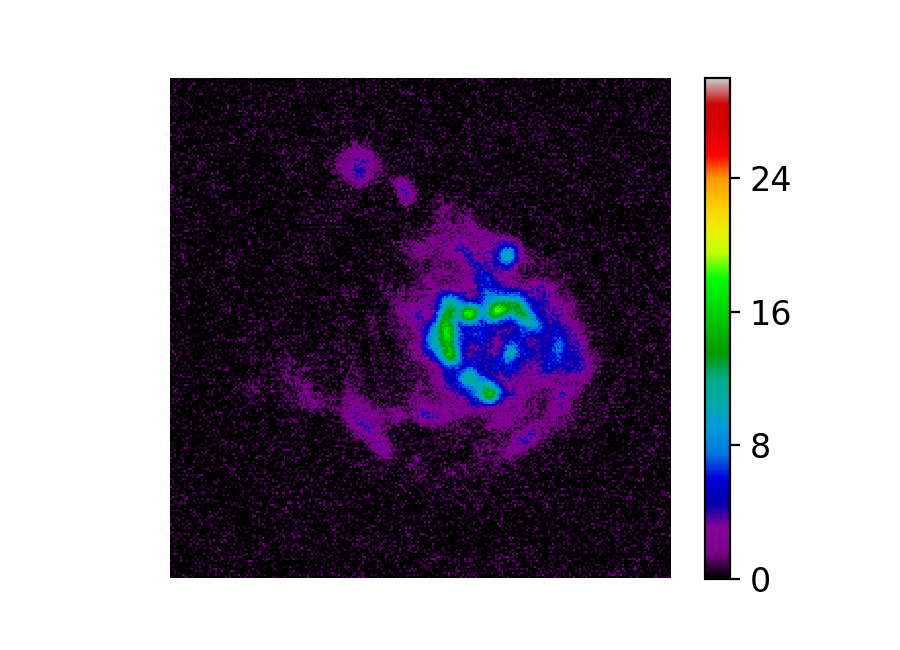} 
  }
\end{minipage}
\begin{minipage}[b]{0.32\linewidth}
  \centering
  \centerline{
   \includegraphics[trim = 2.0cm 0.5cm 1.0cm 0.90cm, clip = true,width=0.99\textwidth,height=0.99\textheight,keepaspectratio,angle=90]{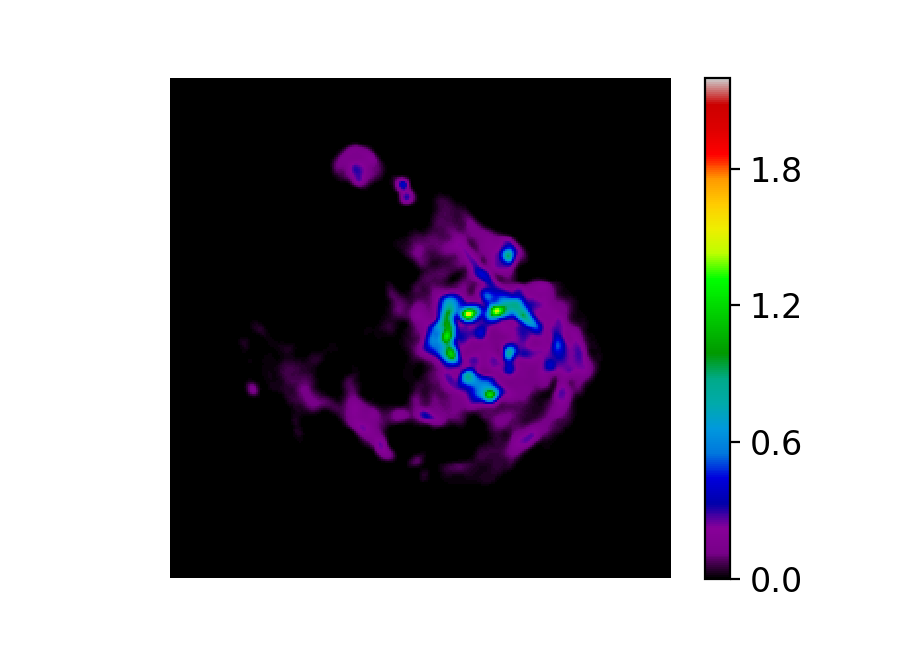}
  }
\end{minipage}
\begin{minipage}[b]{0.32\linewidth}
  \centering
  \centerline{
   \includegraphics[trim = 2.0cm 0.5cm 1.0cm 0.90cm, clip = true,width=0.99\textwidth,height=0.99\textheight,keepaspectratio,angle=90]{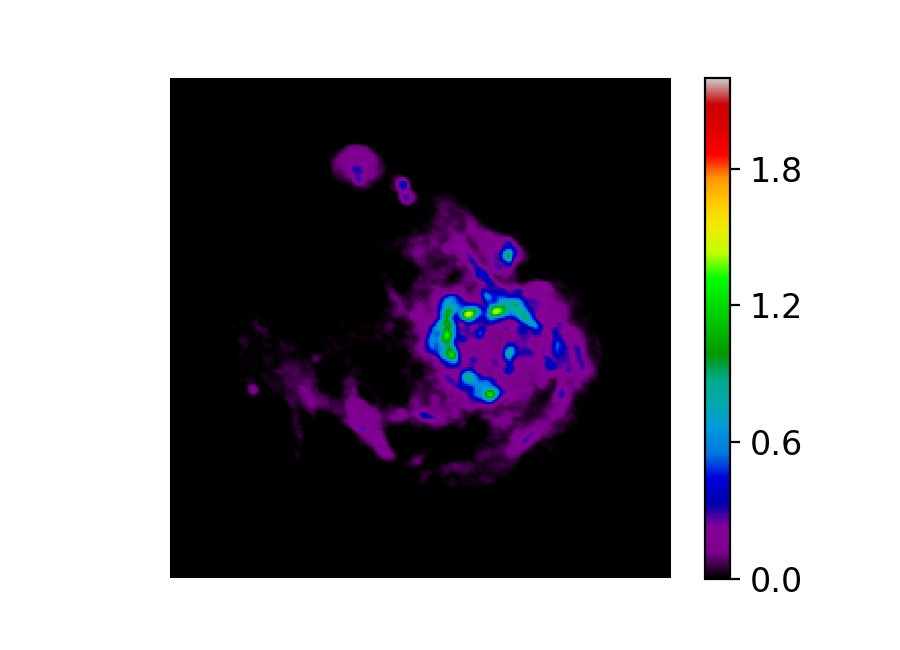}
  }
\end{minipage}


\caption{Left column: M31 Dirty image. Central column: M31 reconstructed images. Right column: M31 sky images.
    First raw, central raw and bottom raw correspond to the initial, central and last wavelength respectively. }
\label{fig:Label}
\end{figure}

\balance
\bibliographystyle{IEEEtran}
\bibliography{MyBiblio}

\end{document}